\documentclass{IEEEtran}
\usepackage{amsmath,amsfonts}
\usepackage{algorithm}
\usepackage{algpseudocode}
\usepackage{array}
\usepackage[caption=false,font=normalsize,labelfont=sf,textfont=sf]{subfig}
\usepackage{textcomp}
\usepackage{stfloats}
\usepackage{url}
\usepackage{verbatim}
\usepackage{graphicx}
\def\BibTeX{{\rm B\kern-.05em{\sc i\kern-.025em b}\kern-.08em
    T\kern-.1667em\lower.7ex\hbox{E}\kern-.125emX}}
\usepackage{balance}
\begin{document}
\title{Enhanced Trust Region Sequential Convex Optimization\\ for Multi-Drone Thermal Screening Trajectory Planning\\ in Urban Environments}
\author{Kaiyuan Chen, \textit{Institute of Automation, Chinese Academy of Science, China} \\Zhengjie Hu, \textit{Beijing Institute of Technology, China} \\Shaolin Zhang, \textit{Institute of Automation, Chinese Academy of Sciences, China} \\Yuanqing Xia, \textit{Beijing Institute of Technology, China}\\ Wannian Liang, \textit{Tsinghua University, China}\\ Shuo Wang, \textit{Institute of Automation, Chinese Academy of Sciences, China}
\thanks{Authors' addresses: 

Kaiyuan Chen is with State Key Laboratory of Multimodal Artificial Intelligence Systems, Institute of Automation, Chinese Academy of Sciences, Beijing 100190, China, also with Vanke School of Public Health, Tsinghua University, Beijing 100084, China;

Zhengjie Hu is with School of Automation, Beijing Institute of Technology, Beijing 100081, China;

Shaolin Zhang is with State Key Laboratory of Multimodal Artificial Intelligence Systems, Institute of Automation, Chinese Academy of Sciences,
Beijing 100190, China;

Yuanqing Xia is with School of Automation, Beijing Institute of Technology, Beijing 100081, China;

Wannian Liang is with Vanke School of Public Health, Tsinghua University, Beijing 100084, China;

Shuo Wang is with State Key Laboratory of Multimodal Artificial Intelligence Systems, Institute of Automation, Chinese Academy of Sciences, Beijing 100190, China, also with the School of Artificial Intelligence,
University of Chinese Academy of Sciences, Beijing 100049, China. 

Corresponding authors: Wannian Liang, liangwn@tsinghua.edu.cn; Shuo Wang, shuo.wang@ia.ac.cn.

This work is founded by National Natural Science Foundation of China. (Grant Number: 72441022)}}

\maketitle

\begin{abstract}
The rapid detection of abnormal body temperatures in urban populations is essential for managing public health risks, especially during outbreaks of infectious diseases. Multi-drone thermal screening systems offer promising solutions for fast, large-scale, and non-intrusive human temperature monitoring. However, trajectory planning for multiple drones in complex urban environments poses significant challenges, including collision avoidance, coverage efficiency, and constrained flight environments. In this study, we propose an enhanced trust region sequential convex optimization (TR-SCO) algorithm for optimal trajectory planning of multiple drones performing thermal screening tasks. Our improved algorithm integrates a refined convex optimization formulation within a trust region framework, effectively balancing trajectory smoothness, obstacle avoidance, altitude constraints, and maximum screening coverage. Simulation results demonstrate that our approach significantly improves trajectory optimality and computational efficiency compared to conventional convex optimization methods. This research provides critical insights and practical contributions toward deploying efficient multi-drone systems for real-time thermal screening in urban areas. For readers who are interested in our research, we release our source code at https://github.com/Cherry0302/Enhanced-TR-SCO.
\end{abstract}

\begin{IEEEkeywords}Multi-Drone; Thermal Screening; Trajectory Planning; Enhanced TR-SCO.
\end{IEEEkeywords}

\section{Introduction}
\IEEEPARstart{R}{apid} urbanization and increasing population densities have amplified the vulnerability of urban centers to infectious disease outbreaks, posing significant challenges to public health systems worldwide \cite{alirol2011urbanisation}. In densely populated cities, the rapid identification and isolation of infected individuals are crucial to mitigating disease transmission and managing public health crises, as clearly demonstrated during recent outbreaks such as COVID-19 and influenza pandemics \cite{lee2020nationwide}.

Thermal screening, a non-contact approach utilizing infrared thermal imaging technology, has emerged as an effective preliminary method for detecting elevated body temperatures—a primary symptom of many infectious diseases—in crowded public spaces such as transportation hubs, shopping centers, schools, and event venues. Unlike traditional temperature measurement methods, thermal screening can be conducted rapidly, safely, and unobtrusively, minimizing discomfort and the risk of cross-infection \cite{perpetuini2021overview,cardwell2021effectiveness}.

However, the effectiveness of manual or stationary thermal screening systems in urban environments is often limited by spatial constraints, low throughput, and challenges in accurately screening large, mobile crowds \cite{khaksari2021review}. Consequently, innovative solutions such as drone-based thermal imaging systems have recently gained attention, offering high mobility, real-time monitoring capability, and adaptability to various urban scenarios \cite{rudisser2021spatially}. Multi-drone systems, in particular, have the potential to significantly enhance coverage and accuracy by simultaneously screening multiple areas or monitoring large crowds dynamically, overcoming limitations associated with ground-based or static thermal screening stations \cite{hoshino2021study}.

Despite these promising advantages, deploying drone-based thermal screening systems introduces unique challenges, particularly regarding trajectory optimization, airspace management, and collision avoidance in complex urban terrains \cite{xu2024application}. Effective trajectory planning algorithms must ensure comprehensive coverage, timely data acquisition, energy efficiency, and adherence to strict aviation safety regulation.

Therefore, developing advanced optimization algorithms capable of handling the complexities inherent in multi-drone coordination and trajectory planning within urban settings is essential. Enhancing such algorithms is expected to significantly improve public health response capabilities, increase the efficiency and accuracy of thermal screening processes, and ultimately contribute to safer and healthier urban communities.

Multi-drone trajectory planning for thermal screening in urban settings involves several complex and interrelated challenges. First, urban environments are characterized by dense infrastructures such as buildings, bridges, and power lines, presenting significant obstacles and constraints for drone navigation. Ensuring effective obstacle avoidance while maintaining efficient flight paths is critical yet challenging due to limited maneuvering space and potential for collision risks \cite{xiang2023study,zhong2024joint}.

Second, multi-drone coordination demands sophisticated methods to avoid inter-drone collisions and ensure optimal area coverage. Drones must dynamically adjust trajectories based on real-time conditions, requiring algorithms capable of rapidly recalculating paths under changing scenarios. The complexity of trajectory synchronization grows exponentially with the number of drones, making real-time trajectory optimization computationally demanding \cite{qazavi2023distributed,du2024model}.

Third, drones performing thermal screening must maintain specific altitudes and orientations to achieve accurate temperature measurements. Inaccurate trajectories or unstable flight patterns can result in measurement errors, reducing the reliability of screening results. Additionally, drones are constrained by limited battery capacity and flight endurance, making energy-efficient trajectory planning crucial for mission success \cite{cho2023comparative,al2022novel}.

Finally, regulatory and safety constraints significantly influence drone operation in urban airspaces. Ensuring compliance with local aviation regulations, privacy laws, and public safety standards adds additional layers of complexity to trajectory planning algorithms. Therefore, addressing these multifaceted challenges requires innovative optimization approaches that integrate safety, accuracy, efficiency, and regulatory compliance within a unified planning framework \cite{pant2021fads,tran2022management}.

Convex optimization methods have been extensively applied to trajectory planning due to their favorable mathematical properties, computational efficiency, and ability to guarantee global optimality under convex constraints \cite{10891025}. The general idea involves formulating trajectory planning as an optimization problem, where an objective function—such as minimizing path length, energy consumption, or trajectory smoothness—is optimized subject to constraints including obstacle avoidance, velocity limits, and boundary conditions \cite{li2022trajectory}.

Common convex optimization approaches include Sequential Convex Programming (SCP), Second-Order Cone Programming (SOCP), and Quadratic Programming (QP). SCP is particularly prominent due to its capability to handle non-convex constraints by iteratively approximating them through convex surrogate problems. Zhang \cite{zhang2022hp} et al. proposed an SCP algorithm based on hp-adaptive Radau pseudospectral discretization (RPD), maintaining solution accuracy and outperforming existing SCP methods and the GPOPS-II tool in overall performance.
Scheffe \cite{scheffe2022sequential} et al. formulated an approximation based on SCP, contrasting conventional sequential linearization (SL) for track constraints, which guaranteed feasibility in the original non-convex problem while preserving recursive feasibility. 
Regarding SOCP, 
Mohammed \cite{el2022hybrid} et al. developed a hybrid robust/stochastic model for transmission expansion planning (TEP), combining stochastic programming (SP) for demand/wind uncertainties and robust optimization (RO) with ellipsoidal uncertainty sets for correlated generator offer prices.
Chowdhury \cite{chowdhury2023second} et al. introduced an improved SOCP-OPF model that enforced the feasibility of the cyclic angle through a convex envelope representation of the relative voltage angles of the bus in the meshed topologies.
In relation to QP, a first-order method for convex quadratic programming named rAPDHG was designed by Lu \cite{lu2023practical} et al. The proposed rAPDHG addressed scalability limitations of traditional simplex/barrier-based solvers reliant on linear equation solving and achieved an optimal linear convergence rate linked to the KKT system’s sharpness constant.
Astudillo \cite{astudillo2022position} et al. applied a simplified sequential convex quadratic programming approach for the predictive control of nonlinear models following the tunnel (NMPC) to address the high complexity of the implementation in existing methods such as generalized Gauss-Newton (GGN). 
Each iteration solves a convexified subproblem, progressively converging toward a feasible and optimal solution of the original non-convex problem.

Recently, Trust Region-based Sequential Convex Optimization (TR-SCO) algorithms have emerged, enhancing the traditional SCP methods by incorporating trust region constraints that limit each optimization iteration to a region near the previous solution. This improves stability, convergence reliability, and computational efficiency, particularly in complex and dynamically changing environments.
Xu \cite{xu2022trust} et al. proposed a trust region filtered sequential convex programming (TRF-SCP) method to enhance computational efficiency in multi-UAV trajectory planning. The approach employed a trust-region filter to dynamically eliminate inactive collision-avoidance constraints in convex subproblems, reducing problem complexity by leveraging spatial relations between trust regions and constraints.
Xie \cite{xie2022oscillation} et al. addressed convergence challenges in the trust-region-based sequential convex programming (TSCP) method for nonlinear aerospace trajectory planning, which suffered from oscillatory behavior despite its real-time suitability.
In order to solve difficulty in convergence due to complex aerodynamic effects, a hybrid-order soft trust region method was proposed by Xie \cite{xie2024hybrid} et al, which combined a low-weight first-order term with higher-order components to balance penalty effects, reformulated subproblems as second-order cone programs (SOCP) via relaxation and integrated line search to ensure convergence. 

Overall, convex optimization approaches provide effective frameworks for trajectory planning due to their structured problem formulation and robust computational properties. However, adapting and enhancing these methods to multi-drone thermal screening scenarios in urban settings necessitates addressing specific challenges, including non-linear environmental constraints, multi-agent coordination, and real-time responsiveness.

In this study, we propose an enhanced Trust Region Sequential Convex Optimization algorithm tailored specifically for trajectory planning of multi-drone thermal screening systems in complex urban environments. Our contributions are threefold: First, we introduce an improved convexification method that effectively addresses the nonlinear and non-convex constraints characteristic of urban drone operations, resulting in enhanced convergence stability and computational efficiency. Second, the integration of adaptive trust-region filtering for constraint reduction and higher-order soft trust-region integration ensures robust trajectory refinement, significantly reducing computation time while maintaining path optimality and safety constraints. Finally, we validate the effectiveness and superiority of our algorithm through comprehensive simulations in realistic urban scenarios, demonstrating marked improvements in thermal screening coverage, trajectory smoothness, and overall system performance compared to conventional convex optimization methods. This research thus represents a significant advancement toward practical and scalable multi-drone thermal screening solutions in urban settings.   

\section{Problem Formulation}

In urban environments, multi-drone thermal screening systems offer a critical solution for rapid, large-scale detection of elevated body temperatures, yet their effectiveness hinges on precise trajectory planning amidst complex constraints like obstacles, inter-drone collisions, and regulatory limits. This section formulates the problem by defining a unified framework that bridges real-world operational requirements—such as safety, efficiency, and coverage—with mathematical modeling. By articulating the system dynamics, optimizing objectives for smooth and efficient flight paths, and enforcing physical, coordination, and regulatory constraints, we lay the groundwork for developing robust algorithms capable of addressing the multifaceted challenges of multi-drone thermal screening in practice.

\subsection{System Model and Notations}

In the context of multi-drone thermal screening trajectory planning, we first establish a comprehensive system model and define a set of notations to precisely describe the problem.

We consider a group of \(N\) drones operating in an urban environment. The trajectory of the \(k\)-th drone in the three-dimensional space is represented as
\begin{equation}
\mathbf{x}_k(t)=[x_k(t), y_k(t), z_k(t)]^T
\end{equation}
where \(t\in\{1,2,\dots,T\}\) is the waypoint number.

The urban environment is modeled as a three-dimensional grid \(\mathcal{M}\in\mathbb{R}^{S\times S\times H}\), where \(S\) represents the horizontal grid size and \(H\) is the range of building heights.

For thermal screening, there are specific constraints. The effective range for temperature detection requires the drones to fly at an appropriate height, typically
\begin{equation}
20\text{m}\leq z_k(t)\leq50\text{m}
\end{equation}
and maintain a specific attitude to ensure the accuracy of infrared imaging.

We also define several important notations. The set of obstacle positions is denoted as \(\mathcal{O}\), and the safety distance is represented by \(d_{\text{safe}}\). 

This system model and the associated notations form the basis for formulating the subsequent objective functions and constraints in the multi-drone thermal screening trajectory planning problem.

\subsection{Objective Function Formulation}

In the multi-drone thermal screening trajectory planning problem, we aim to formulate an objective function that takes into account both the smoothness of the drone trajectories and the efficiency of the flight paths.

The smoothness of the drone trajectories is crucial to avoid sudden and drastic maneuvers of the drones. We measure the smoothness of the trajectory of the \(k\)-th drone over the waypoints from \(t = 2\) to \(t=T - 1\) using the following formula:
\begin{equation}
J_{\text{smooth}}=\sum_{t = 2}^{T-1}\left\|\mathbf{x}_k(t + 1)-2\mathbf{x}_k(t)+\mathbf{x}_k(t - 1)\right\|_2
\end{equation}
where \(\mathbf{x}_k(t)=[x_k(t),y_k(t),z_k(t)]^T\) represents the position of the \(k\)-th drone at waypoint \(t\) in the three-dimensional space, and \(\|\cdot\|_2\) is the Euclidean norm.

The efficiency of the flight paths is measured by minimizing the total flight distance of the drones. For the \(k\)-th drone, the total flight distance over the waypoints from \(t = 1\) to \(t=T - 1\) is calculated as:
\begin{equation}
J_{\text{length}}=\sum_{t = 1}^{T-1}\left\|\mathbf{x}_k(t + 1)-\mathbf{x}_k(t)\right\|_2
\end{equation}

The overall objective function \(J\) is the combination of the smoothness and the path length, which can be expressed as:
\begin{equation}
J=J_{\text{smooth}}+J_{\text{length}}
\end{equation}

This objective function provides a balanced approach to optimize the drone trajectories in terms of both smoothness and path efficiency, ensuring that the drones can perform the thermal screening tasks in a stable and efficient manner.

\subsection{Constraints Definition}

In the multi-drone thermal screening trajectory planning problem, a set of constraints must be defined to ensure the safety, feasibility, and effectiveness of the drone operations.

\subsubsection{Physical and Safety Constraints}

The first set of constraints is related to the physical limitations and safety requirements of the drones. To avoid collisions with buildings and other obstacles, the drones must maintain a safe distance from them. Let \(h_{\text{obs}}(x_k(t), y_k(t))\) be the height function of the obstacles at the horizontal position \((x_k(t), y_k(t))\) of the \(k\)-th drone at waypoint \(t\). The obstacle-avoidance constraint can be expressed as
\begin{equation}
z_k(t)\geq h_{\text{obs}}(x_k(t), y_k(t)) + d_{\text{safe}}
\end{equation}
where \(d_{\text{safe}}\) is the predefined safety distance.

The drones are also required to stay within the boundaries of the simulation area. This can be formulated as
\begin{equation}
0\leq x_k(t), y_k(t)\leq S - 1
\end{equation}
where \(S\) is the horizontal grid size of the urban environment model.

Moreover, the drones must fly within a certain height range to ensure effective thermal screening. The height constraint is given by
\begin{equation}
z_{\text{min}}\leq z_k(t)\leq z_{\text{max}}
\end{equation}
where \(z_{\text{min}}\) and \(z_{\text{max}}\) are the minimum and maximum allowable flight heights, respectively.

Additionally, in order to ensure that the drones approach the preset waypoints $(x(t),y(t)$, it is formulated that
\begin{equation}
    \sqrt{((x_k(t)-x(t))^2+(y_k(t)-y(t))^2)}\leq d_{\text{dev}}
\end{equation}
where \(d_{\text{dev}}\) is the maximum deviation distance.

\subsubsection{Multi-Drone Coordination Constraints}

When multiple drones are operating in the same area, coordination is necessary to prevent collisions between them. The distance between any two drones \(k\) and \(m\) at waypoint \(t\) must be greater than or equal to a predefined safety distance \(d_{\text{drone}}\). This can be written as
\begin{equation}
\left\|\mathbf{x}_k(t)-\mathbf{x}_m(t)\right\|_2\geq d_{\text{drone}}, \quad k\neq m
\end{equation}

In addition, to ensure efficient task allocation, the drones should cover the screening area without overlap or redundancy. This can be achieved through a waypoint allocation matrix \(\mathcal{W}_k\) that divides the screening area among the drones.

\subsubsection{Regulations and Performance Constraints}

The drone operations must comply with urban airspace flight rules, such as no-fly zones. Also, considering the battery life of the drones, the total length of the flight path of each drone should not exceed the maximum flight distance allowed by its battery capacity.

These constraints, along with the objective function defined in the previous section, form a complete mathematical model for the multi-drone thermal screening trajectory planning problem.

\section{Methodology}

\subsection{Introduction to the Trust Region Sequential Convex Optimization Algorithm}

In the trajectory optimization problem under complex constraints, the trust region sequential convex optimization algorithm effectively balances the local exploration and global convergence of the optimization process by combining the sequential convex programming framework with the trust region strategy. The core idea of this method is to iteratively solve a series of convex sub-problems to gradually approach the optimal solution of the original nonlinear problem. Meanwhile, the trust region constraint is used to limit the search range of each iteration, avoiding instability caused by an overly large step size.

Consider the state equation of a nonlinear dynamic system:
\begin{equation}
\dot{\mathbf{x}}(t) = \mathbf{f}(\mathbf{x}(t), \mathbf{u}(t)),
\end{equation}
where $\mathbf{x} \in \mathbb{R}^n$ is the state vector, $\mathbf{u} \in \mathbb{R}^m$ is the control input, and $\mathbf{f}$ is a nonlinear vector function. In the TR-SCO framework, the current reference trajectory $(\mathbf{x}^r(t), \mathbf{u}^r(t))$ is first linearized. By retaining the first-order terms through Taylor expansion, a linearized dynamic model is obtained:
\begin{equation}
\dot{\mathbf{x}}(t) \approx \mathbf{A}(t)\mathbf{x}(t) + \mathbf{B}(t)\mathbf{u}(t) + \mathbf{c}(t),
\end{equation}
where
\begin{equation}
\mathbf{A}(t) = \frac{\partial \mathbf{f}}{\partial \mathbf{x}}\big|_{(\mathbf{x}^r, \mathbf{u}^r)}
\end{equation}
and 
\begin{equation}
    \mathbf{B}(t) = \frac{\partial \mathbf{f}}{\partial \mathbf{u}}\big|_{(\mathbf{x}^r, \mathbf{u}^r)}
\end{equation}
are the Jacobian matrices of the state and control, and 
\begin{equation}
    \mathbf{c}(t) = \mathbf{f}(\mathbf{x}^r(t), \mathbf{u}^r(t)) - \mathbf{A}(t)\mathbf{x}^r(t) - \mathbf{B}(t)\mathbf{u}^r(t)
\end{equation}
is a constant term.

For the nonlinear constraint $\mathbf{g}(\mathbf{x}(t), \mathbf{u}(t)) \leq \mathbf{0}$, a first-order Taylor approximation is also used to transform it into a convex constraint:
\begin{equation}
\begin{aligned}
\mathbf{g}(\mathbf{x}^r(t), \mathbf{u}^r(t))+\nabla_{\mathbf{x}}\mathbf{g}^T\big|_{(\mathbf{x}^r, \mathbf{u}^r)}(\mathbf{x}(t)-\mathbf{x}^r(t))+\\
\nabla_{\mathbf{u}}\mathbf{g}^T\big|_{(\mathbf{x}^r, \mathbf{u}^r)}(\mathbf{u}(t)-\mathbf{u}^r(t)) \leq \mathbf{0}.
\end{aligned}
\end{equation}
To ensure that the iterative process is carried out within a reliable region, a trust region constraint is introduced:
\begin{equation}
\|\mathbf{x}(t)-\mathbf{x}^r(t)\|_2 \leq \delta_x, \quad \|\mathbf{u}(t)-\mathbf{u}^r(t)\|_2 \leq \delta_u,
\end{equation}
where $\delta_x$ and $\delta_u$ are the trust region radii, which limit the variation ranges of the state and control variables, avoiding infeasible solutions caused by excessive linearization approximation errors.

By transforming the original problem into an optimization sub-problem based on the linearized dynamics and convex constraints, and adding a trajectory smoothing term and a path length term to the objective function, the TR-SCO algorithm can solve a convex optimization problem at each iteration step and gradually update the reference trajectory until convergence. This strategy not only takes advantage of the efficient solution characteristics of convex optimization but also enhances the robustness of the algorithm through the trust region mechanism. It is particularly suitable for dealing with the complex nonlinear constraints of multi-drone trajectory planning in urban environments, such as obstacle avoidance, collision avoidance between drones, and flight altitude limits.

\subsection{Enhancements for TR-SCO}
\subsubsection{Convexification}
To address the unique challenges of multi-drone thermal screening in urban environments, the proposed TR-SCO algorithm incorporates specific enhancements tailored to the problem's constraints. Central to this is the convexification of the nonlinear and non-convex constraints defined in Section 2.3, enabling efficient solution within the trust region framework.

The obstacle avoidance constraint, originally formulated as \( z_k(t) \geq h_{\text{obs}}(x_k(t), y_k(t)) + d_{\text{safe}} \), is non-convex due to the nonlinear height function \( h_{\text{obs}} \). To convexify this, we approximate \( h_{\text{obs}} \) around the current reference trajectory \( (x_k^r(t), y_k^r(t)) \) using a first-order Taylor expansion. This yields the linearized constraint:
\begin{equation}
z_k(t) \geq h_{\text{obs}}(x_k^r(t), y_k^r(t)) + \nabla_{x,y} h_{\text{obs}}^T \begin{bmatrix} x_k(t) - x_k^r(t) \\ y_k(t) - y_k^r(t) \end{bmatrix} + d_{\text{safe}},
\end{equation}
where \( \nabla_{x,y} h_{\text{obs}} \) denotes the gradient of the obstacle height function at the reference point. This linear approximation is valid within the trust region radius \( \delta_{x,y} \), ensuring the constraint remains convex and tractable. 
The flight altitude constraint \( z_{\text{min}} \leq z_k(t) \leq z_{\text{max}} \) and boundary constraints \( 0 \leq x_k(t), y_k(t) \leq S - 1 \) are inherently linear and thus directly incorporated as convex constraints without modification.

For waypoint adherence, the original nonlinear deviation constraint \(\sqrt{(x_k(t) - x(t))^2 + (y_k(t) - y(t))^2} \leq d_{\text{dev}}\) is linearized into component-wise absolute value constraints to maintain convexity value constraints to maintain convexity:
\begin{equation}
|x_k(t) - x(t)| \leq d_{\text{dev}}, \quad |y_k(t) - y(t)| \leq d_{\text{dev}}.
\end{equation}
This ensures drones remain within a safe positional margin for accurate thermal imaging, critical for maintaining the effective detection range enforced by the height constraint.

To address the complex constraints and dynamic requirements of multi-drone thermal screening in urban environments, we introduce two key enhancements to the traditional TR-SCO framework, integrating insights from trust-region filtering and higher-order optimization strategies.
\subsubsection{Adaptive Trust-Region Filtering for Constraint Reduction}
Inspired by the trust-region filtered SCP (TRF-SCP) approach, we develop a dynamic constraint filtering mechanism to reduce computational complexity. Urban environments impose numerous collision-avoidance constraints (e.g., building obstacles and inter-drone spacing), but many constraints are inactive during specific flight segments. The filtering mechanism operates as follows:
1)Obstacle Avoidance Filter: For each drone's reference trajectory \(x_k^r(t)\), we define a trust region
\begin{equation}
 \|x_k(t) - x_k^r(t)\|_2 \leq \delta \ 
\end{equation}
and check if obstacles lie within this region. Obstacles outside the trust region are excluded from the current subproblem.
2)Inter-Drone Collision Filter: Similarly, we filter inter-drone collision constraints by checking if the distance between any two drones \( \|x_k(t) - x_m(t)\|_2 \) falls within the trust region. Only constraints that intersect the trust region are retained, significantly reducing the number of active constraints.

This filtering strategy reduces the convex subproblem size, enabling real-time optimization for large drone fleets.
To balance exploration and convergence stability, we introduce an adaptive trust-region updating mechanism. The radius \( \delta \) is adjusted iteratively based on constraint violation. If the constraint violation \( \ell(X^q) \) exceeds a threshold \( \varepsilon_\rho \), the radius expands
\begin{equation}
 \delta^{q+1} = c_2 \cdot \delta^q ,  c_2 > 1 
\end{equation}
to broaden the search space and resolve infeasibility. If \( \ell(X^q) \leq \varepsilon_\rho \), the radius contracts
\begin{equation}
 \delta^{q+1} = c_1 \cdot \delta^q ,  0 < c_1 < 1
\end{equation}
to refine the solution and reduce linearization errors.

This mechanism dynamically eliminates inactive collision-avoidance constraints that lie outside the trust region, reducing the convex subproblem size. By focusing only on active constraints, the algorithm achieves faster computation compared to traditional SCP methods, making it feasible for real-time optimization in large drone fleets. This is critical for urban scenarios where dynamic obstacle avoidance and rapid trajectory updates are essential.
\subsubsection{Higher-Order Soft Trust-Region Integration}
To enhance solution optimality, we design a higher-order soft trust region, transforming the trust-region constraint from a first-order penalty to a higher-order term (e.g., \( \delta^p \), \( p > 1 \)) in the objective function. This modification:
1)Guarantees KKT optimality: Higher-order terms (e.g., \( p=2 \)) ensure that the converged solution satisfies the Karush–Kuhn–Tucker (KKT) conditions, whereas first-order terms may lead to suboptimal solutions. 2)Suppresses oscillation: The higher-order penalty penalizes large trust-region radii more strongly, stabilizing the iteration process and reducing fluctuations in trajectory smoothness.

The objective function is adjusted to:

\begin{equation}
\min J = J_{\text{smooth}} + J_{\text{length}} + w_\delta \cdot \delta^p
\end{equation}
where \( w_\delta \) is the penalty weight and \( p \geq 2 \) is the order. This ensures that thermal screening trajectories are both optimal and smooth, critical for maintaining infrared imaging accuracy.

Higher-order penalties enforce smaller trust-region radii near convergence, reducing jerk and improving thermal imaging stability. This is critical for maintaining infrared camera focus, as abrupt movements can introduce measurement errors.


\subsection{Mathematical Formulation and Optimization Approach}

The mathematical formulation of the enhanced Trust Region Sequential Convex Optimization algorithm integrates the core objectives of multi-drone thermal screening—smooth trajectories and efficient flight paths with a structured set of constraints within a unified optimization framework. The objective function balances two key goals: minimizing abrupt changes in drone motion to ensure stable thermal imaging and reducing total flight distance to conserve energy. The mathematical formulation is as follows:
\begin{equation}
\begin{aligned}
\min \quad & J = J_{\text{smooth}} + J_{\text{length}}+ w_\delta \cdot \delta^p\\
\text{s.t.} \quad & z_k(t) \geq h_{\text{obs}}(x_k^r(t), y_k^r(t)) +\\& \nabla_{x,y} h_{\text{obs}}^T \begin{bmatrix} x_k(t) - x_k^r(t) \\ y_k(t) - y_k^r(t) \end{bmatrix} + d_{\text{safe}}, \quad \forall t \\
& \|\mathbf{x}_k(t) - \mathbf{x}_m(t)\|_2 \geq d_{\text{drone}}, \quad \forall k \neq m, \forall t \\
& z_{\text{min}} \leq z_k(t) \leq z_{\text{max}}, \quad \forall t \\
& 0 \leq x_k(t), y_k(t) \leq S - 1, \quad \forall t \\
&|x_k(t) - x(t)| \leq d_{\text{dev}}, \quad |y_k(t) - y(t)| \leq d_{\text{dev}}, \quad \forall t
\end{aligned}
\end{equation}
where each variable is clearly stated in the previous sections.

The optimization process adheres to multiple constraints designed to ensure safety, coordination, and regulatory compliance. Physical and safety constraints include maintaining a safe distance above obstacles (approximated using local terrain data around the current flight path), staying within specified altitude limits for accurate thermal detection, and operating within the boundaries of the urban area model. Multi-drone coordination constraints enforce safe distances between individual drones to prevent collisions and synchronize their speeds to maintain coherent group motion, avoiding both dangerous proximity and inefficient redundant coverage. Regulatory and performance constraints incorporate no-fly zones and battery limitations, ensuring each drone’s total flight distance does not exceed its operational capacity.

A critical component of the framework is the adaptive trust region strategy, which limits each iteration’s search space to a neighborhood around the current reference trajectory. This mechanism ensures that linear approximations of complex nonlinear constraints—such as irregular obstacle heights or dynamic drone interactions—remain valid and stable, avoiding large, infeasible adjustments that could violate safety margins. The size of the trust region is dynamically adjusted based on the algorithm’s progress: it contracts when constraints are frequently violated to encourage more precise local refinement and expands as the solution converges to explore broader adjustments.

The enhanced TR-SCO algorithm iteratively solves a series of convex subproblems to approach the optimal solution of the original nonlinear problem. The core idea is to linearize the nonlinear system around the current reference trajectory and solve a convex optimization problem at each iteration, while limiting the search range using trust region constraints. The proposed enhanced TR-SCO algorithm can be described with the following steps presented in Algorithm 1.
\begin{algorithm}
\caption{Enhanced Trust Region Sequential Convex Optimization Algorithm}
\begin{algorithmic}[1]
\State \textbf{Initialize:} Set $\mathbf{x}_{\text{ref}}(t) = \mathbf{x}_0(t)$, iteration counter $iter = 0$
\Repeat
    \State Linearize system dynamics and constraints around $\mathbf{x}_{\text{ref}}(t)$:
    \State Use first-order Taylor expansion for nonlinear constraints
    \begin{equation}
    \begin{aligned}
        z_k(t) \geq h_{\text{obs}}(x_k^r(t), y_k^r(t)) +\\ \nabla_{x,y} h_{\text{obs}}^T \begin{bmatrix} x_k(t) - x_k^r(t) \\y_k(t) - y_k^r(t) \end{bmatrix} + d_{\text{safe}}
        \end{aligned}
    \end{equation}
    \State Formulate the convex subproblem with trust region constraints:
    \begin{equation}
    \begin{aligned}
    \min \quad & J=J_{\text{smooth}} + J_{\text{length}}+ w_\delta \cdot \delta^p\\
    \text{s.t.} \quad & z_k(t) \geq h_{\text{obs}}(x_k^r(t), y_k^r(t)) + \\&\nabla_{x,y} h_{\text{obs}}^T \begin{bmatrix} x_k(t) - x_k^r(t) \\ y_k(t) - y_k^r(t) \end{bmatrix} + d_{\text{safe}}\\
    & \|x_k(t) - x_m(t)\|_2 \geq d_{\text{drone}} \\
    & z_{\text{min}} \leq z_k(t) \leq z_{\text{max}} \\
    & 0 \leq x_k(t), y_k(t) \leq S - 1 \\
    &|x_k(t) - x(t)| \leq d_{\text{dev}}, \quad |y_k(t) - y(t)| \leq d_{\text{dev}}
    \end{aligned}
    \end{equation}
    \State Solve the convex subproblem to obtain \( \mathbf{x}_{\text{new}}(t) \)
    \State Update the reference trajectory: \( \mathbf{x}_{\text{ref}}(t) = \mathbf{x}_{\text{new}}(t) \)
    \State Update trust region radius
    \begin{equation}
    \begin{cases} 
\delta^{q+1} = c_2 \delta^q, & c_2 > 1, \quad \text{if } \ell(X^q) > \varepsilon_\rho, \\ 
\delta^{q+1} = c_1 \delta^q, & 0 < c_1 < 1, \quad \text{if } \ell(X^q) \leq \varepsilon_\rho 
\end{cases}
\end{equation}
    \State Increment iteration counter: \( iter = iter + 1 \)
\Until{Convergence criterion is met}
\State \textbf{Output:} Optimized trajectory
\end{algorithmic}
\end{algorithm} 

Algorithm 1 effectively balances local exploration and global convergence by iteratively solving convex subproblems within a trust region framework. The trust region constraints ensure that each iteration remains within a reliable region around the previous solution, avoiding instability caused by overly large step sizes. Through this approach, the enhanced TR-SCO algorithm efficiently handles the complex nonlinear constraints of multi-drone trajectory planning in urban environments, ensuring safety, coverage, and computational efficiency for thermal screening missions.
 
\section{Simulation and Results}

\subsection{Description of Simulation Environment and City Model}
In this study, we developed a realistic simulation environment to evaluate the performance of our enhanced Trust Region Sequential Convex Optimization algorithm for multi-drone thermal screening missions. The simulation was implemented in Python using widely-adopted scientific and visualization libraries, including numpy, cvxpy, and matplotlib.

The urban environment was modeled as a grid-based, three-dimensional cityscape, represented by a
50×50 discretized area. Buildings and static obstacles within this environment were randomly generated, with a controlled density of approximately 20–30\%. Each building was modeled as a rectangular prism (cuboid), with randomly assigned heights varying uniformly between 10 m and 50 m. This approach effectively replicates typical urban complexity, allowing the testing of drone navigation and obstacle avoidance capabilities within densely built-up areas.

Multi-drone operations were simulated using five drones, each assigned randomized initial positions within obstacle-free areas of the city model. Drone trajectories were defined by a sequence of waypoints, optimized using our enhanced TR-SCO method. The generated trajectories adhered to several operational constraints, including safe margin distances from buildings (5 m), allowable flight altitude ranges (20–50 m), and boundary conditions within the simulated area.

\subsection{Results and Analysis}



To systematically assess the performance of the enhanced Trust Region Sequential Convex Optimization algorithm, we evaluated three key metrics: path efficiency (total path length, lower values indicate better energy efficiency), thermal coverage area (total coverage area, lower values indicate more comprehensive screening, balanced against redundancy), and calculation time (time to converge, lower values indicate better real-time applicability). Comparisons were made with three methods: The Original TR-SCO Algorithm(Original TR-SCO, TR-SCO without being enhanced), Mixed-Integer Nonlinear Programming (MINLP, a non-convex optimization method involving integer variables and nonlinear functions) \cite{androulakis2024minlp} and SCS (Splitting Conic Solver, a conventional convex optimization approach for solving large-scale convex cone problems) \cite{chen2024efficient}.

\begin{figure}
    \centering
    \includegraphics[width=0.9\linewidth]{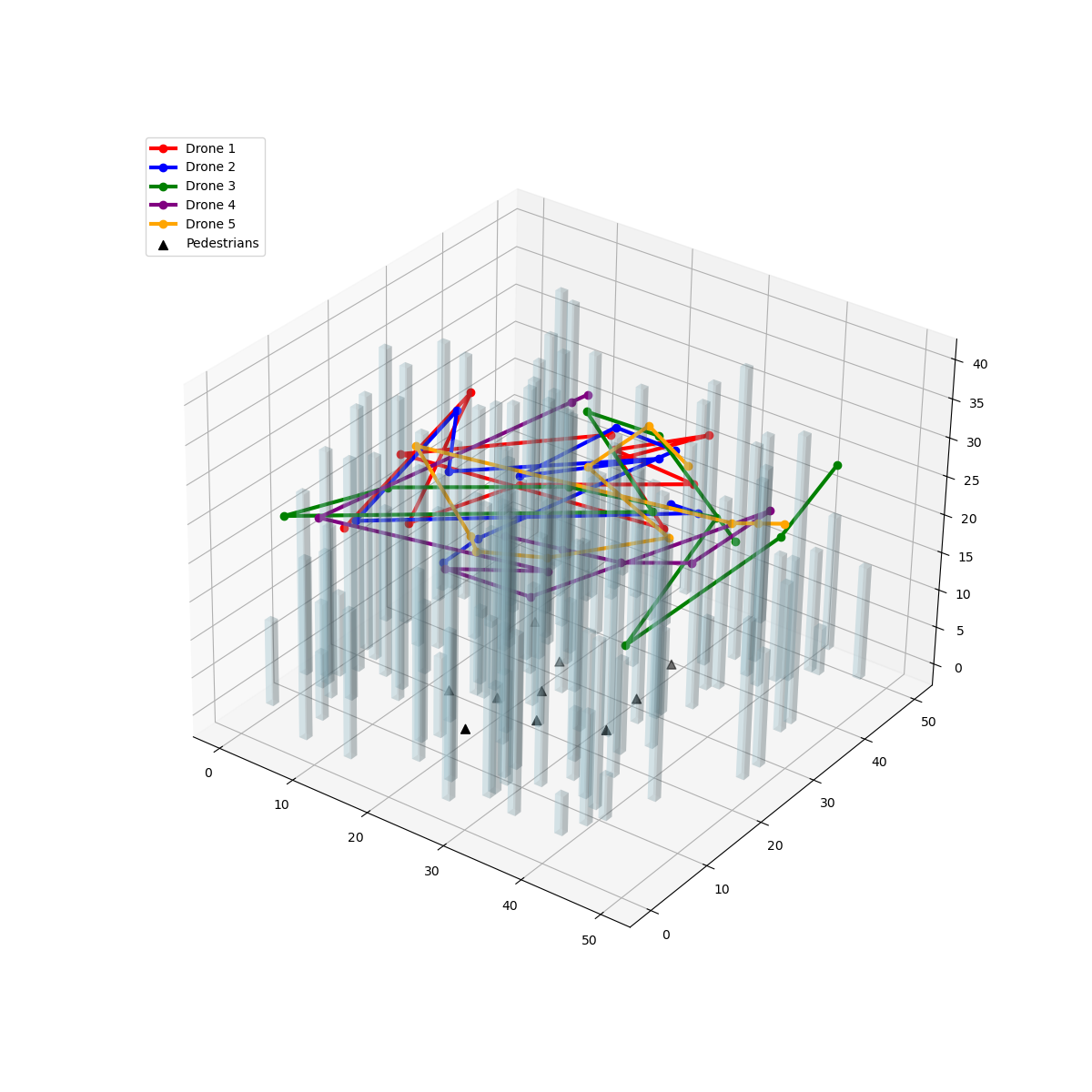}
    \caption{Trajectories of 5-drone Scenario 1}
    \label{fig:5 drones}
\end{figure}

Figure \ref{fig:5 drones} illustrates the optimized trajectories of five drones navigating a simulated urban environment designed for thermal screening. The environment, represented as a 50$\times$50 3D grid with 20-30\% building density, imposes constraints such as a 20-50 m altitude range for effective infrared temperature detection and safe distances from obstacles. The trajectories demonstrate the enhanced TR-SCO algorithm's ability to generate smooth, energy-efficient paths that avoid abrupt maneuvers, critical for maintaining stable drone orientation and accurate thermal imaging. Drones navigate around buildings and other obstacles while maintaining safe inter-drone spacing, showcasing the algorithm's integration of convexified nonlinear constraints—like obstacle height approximations via Taylor expansion within trust regions—to ensure safety and feasibility. The paths prioritize coverage of pedestrian-dense areas, reflecting the algorithm's dynamic waypoint allocation to minimize redundant screening and maximize efficiency in crowded zones, a key requirement for effective public health monitoring.

\begin{figure}
    \centering
    \includegraphics[width=0.9\linewidth]{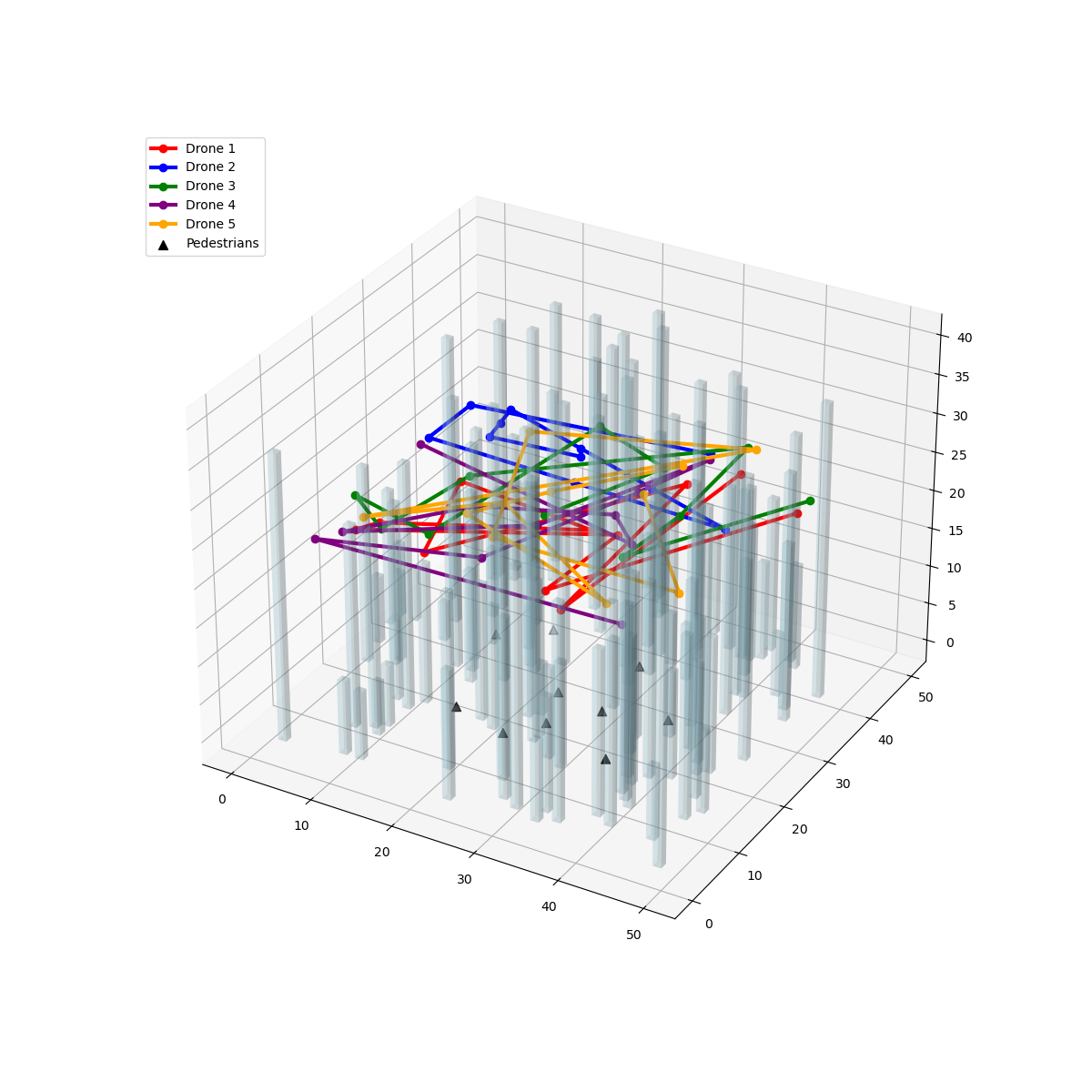}
    \caption{Trajectories of 5-drone Scenario 2}
    \label{fig:5 drones2}
\end{figure}

Depicting a different urban configuration with the same environmental parameters but potentially varied obstacle layouts or initial drone positions, Figure \ref{fig:5 drones2} highlights the algorithm's adaptability to diverse urban scenarios. The trajectories remain smooth and collision-free, demonstrating the robustness of the trust region strategy that dynamically adjusts the search space to balance local refinement and global convergence. Drones maintain uniform safe distances from one another, verifying the effectiveness of modeling inter-drone collision constraints as convex second-order cone problems, which avoids the inefficiencies or instability seen in baseline methods. All operational constraints are strictly met: drones fly within the specified altitude range and grid boundaries, integrating physical, safety, and regulatory limits into a unified optimization framework. The coordinated motion across varying environments underscores the algorithm's capacity to handle complex, real-world urban dynamics, ensuring reliable performance even as obstacle distributions or mission requirements change.
\begin{table}
\centering
\caption{Performance Comparison for 5 Drones}
\label{tab:5 drones}
\scalebox{0.67}{
\begin{tabular}{cccc}
\hline
\textbf{Method} & \textbf{Total Path Length(m)} & \textbf{Total Coverage Area(m$^2$)} & \textbf{Calculation Time(s)} \\ \hline
Enhanced TR-SCO       & \textbf{859.11}            & \textbf{2490.74}             & 0.6511               \\
Original TR-SCO &916.14&2574.62&0.8621\\
MINLP        & 986.89            & 2783.25             & \textbf{0.5495}               \\
SCS          & 956.42            & 2812.85             & 4.6958  \\ \hline
\end{tabular}
}
\end{table}

Table \ref{tab:5 drones} presents a performance comparison of four trajectory planning methods for 5 drones in urban thermal screening, evaluating total path length, total coverage area, and calculation time. The enhanced TR-SCO algorithm achieves the shortest total path length at 859.11 meters, demonstrating superior energy efficiency by balancing obstacle avoidance, inter-drone spacing, and dynamic trust region adjustments to minimize detours. In contrast, the Original TR-SCO, MINLP, and SCS show longer paths due to less refined constraint handling: the original lacks terrain-adaptive safety margins, MINLP struggles with local minima in discrete optimization, and SCS suffers from unstable linearization without trust region bounds.

In the 5-drone scenario, the enhanced TR-SCO algorithm achieves a 12.9\% shorter total path length compared to MINLP and a 10.2\% shorter path than SCS. This efficiency stems from its ability to balance obstacle avoidance, inter-drone spacing, and coverage optimization without excessive detours. The smallest coverage area indicates minimal redundant screening, a critical benefit for focusing on high-pedestrian density zones (see Figure \ref{fig:5 drones}). 

The original TR-SCO algorithm underperforms due to less adaptive constraint handling, static trust region strategies, and insufficient multi-drone coordination. It lacks the enhanced version’s terrain-adaptive safety margins and dynamic trust region radius adjustment, leading to suboptimal obstacle avoidance and longer paths. Without velocity synchronization constraints, it struggles with coherent drone motion, increasing coverage redundancy. The enhanced algorithm’s refined convexification of nonlinear constraints and mission-tailored objective function further improve efficiency, while its adaptive trust region reduces computation time by ensuring stable linear approximations. Simulation results confirm the original’s longer paths, larger coverage areas, and slower convergence relative to the enhanced TR-SCO algorithm, which integrates these advancements for safer, more precise urban trajectory planning.

While MINLP can handle non-convex constraints, it suffers from combinatorial explosion. Its reliance on integer variables for discrete decisions and nonlinear solvers for collision avoidance often leads to suboptimal local minima, especially in densely built environments. Although its computational time appears deceptively fast, it sacrifices solution quality—its coverage area is 14.9\% larger than the enhanced TR-SCO algorithm. Computational efficiency is a standout advantage of the enhanced TR-SCO algorithm. It completes optimization in 0.65 seconds, significantly faster than SCS and marginally slower than MINLP. This balance of optimality and speed is critical for real-time urban operations.

As a conventional convex optimization method, SCS lacks the adaptive trust region mechanism of the enhanced TR-SCO algorithm. While it iteratively linearizes non-convex constraints, it does not dynamically restrict the search space, leading to two key issues. One is unstable linearization approximations. Without trust region bounds, large step sizes in early iterations often violate obstacle or inter-drone distance constraints, requiring more iterations for correction. The other is suboptimal constraint convexification. SCS uses first-order Taylor expansions for all constraints but does not integrate second-order cone programming for inter-drone distance constraints, resulting in loose approximations and redundant safety margins. This explains why SCS’s coverage area is 13.0\% larger than the enhanced TR-SCO algorithm.
\begin{figure}
    \centering
    \includegraphics[width=0.9\linewidth]{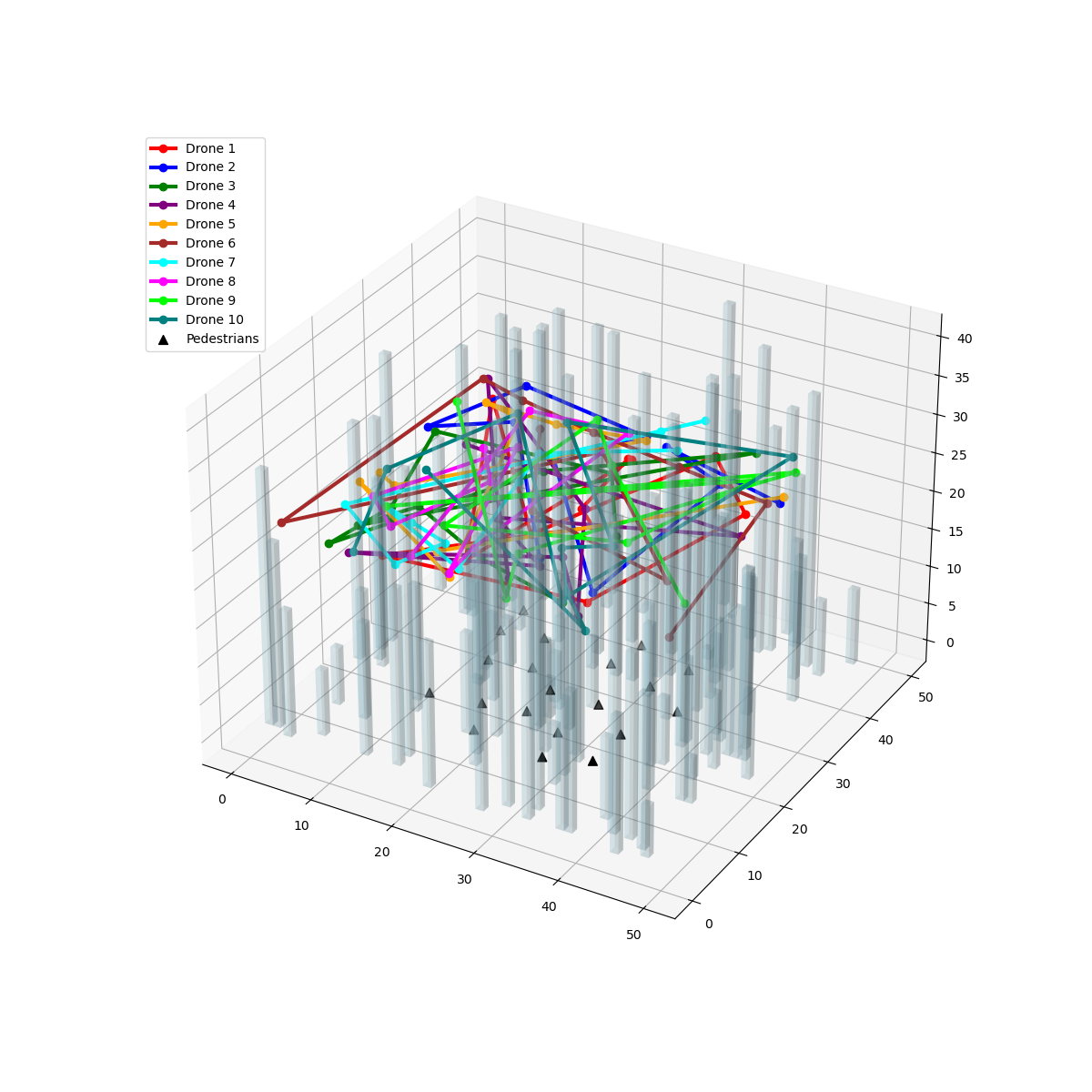}
    \caption{Trajectories of 10-drone Scenario 1}
    \label{fig:10 drones}
\end{figure}

\begin{figure}
    \centering
    \includegraphics[width=0.9\linewidth]{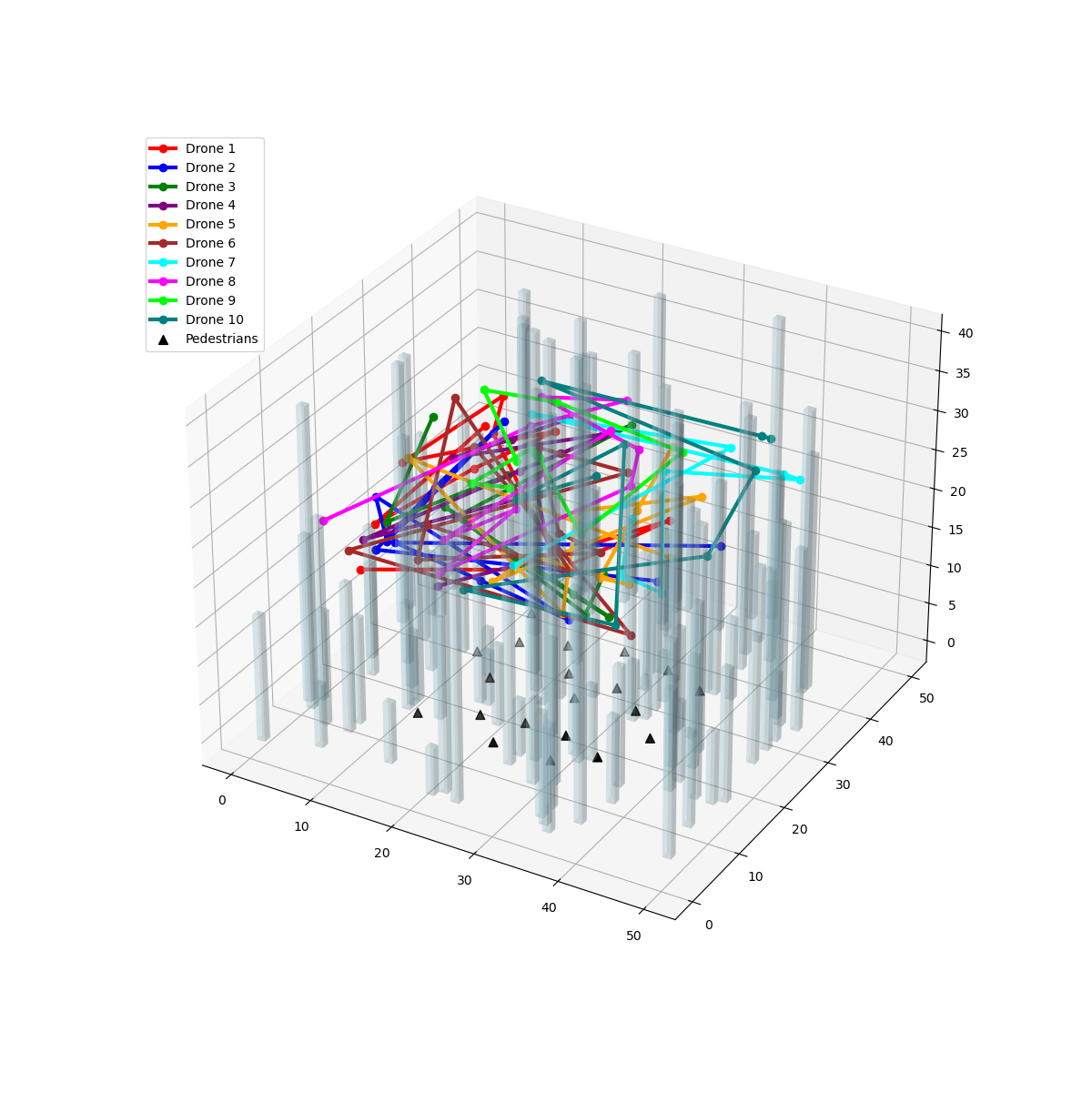}
    \caption{Trajectories of 10-drone Scenario 2}
    \label{fig:10 drones2}
\end{figure}

As the number of drones increases to 10, Figure \ref{fig:10 drones} presents the optimized trajectories of 10 drones in an urban grid, avoiding obstacles with smooth trajectories while maintaining safe distances between drones. Trajectories prioritize high-pedestrian areas for efficient thermal screening, demonstrating the balance of safety and coverage of the algorithm, while Figure \ref{fig:10 drones2} depicts adaptability in a different urban layout, with drones navigating smoothly within altitude/grid constraints. Even with 10 drones, the proposed algorithm ensures collision-free coordination and rapid convergence, proving scalability for complex multi-drone missions. Both figures validate the ability and adaptability of the enhanced TR-SCO algorithm to generate safe, efficient trajectories for large drone fleets in urban environments, critical for effective thermal screening in dense populations.

Table \ref{tab:TR-SCO 5 and 10} shows a proportional increase in total path length and coverage area, reflecting the need for more complex coordination. Despite the higher problem complexity, the enhanced TR-SCO algorithm maintains computational efficiency with a convergence time of 1.58 seconds, demonstrating scalability.

\begin{table}
\centering
\caption{The Enhanced TR-SCO Algorithm Performance for 5 and 10 Drones}
\label{tab:TR-SCO 5 and 10}
\scalebox{0.7}{
\begin{tabular}{cccc}
\hline
\textbf{Drone Number} & \textbf{Total Path Length(m)} & \textbf{Total Coverage Area(m$^2$)} & \textbf{Calculation Time(s)} \\ \hline
5 & 859.11 & 2490.74 & 0.6511 \\
        10 & 1684.91 & 4934.46 & 1.5751 \\
 \hline
\end{tabular}
}
\end{table}

The enhanced TR-SCO algorithm restricts each iteration’s solution within a local neighborhood, ensuring linearized approximations of nonlinear constraints remain valid and avoiding the step-size instability that plagues SCS. Inter-drone collision constraints are transformed, which enables precise distance control by convex solvers without over-approximation. Non-convex obstacle avoidance constraints, caused by nonlinear building heights, are linearized via first-order Taylor expansions around reference trajectories within trust regions, ensuring safe navigation through complex urban geometries with minimal detour (as seen in Figure \ref{fig:10 drones}, where drones smoothly traverse gaps between tall buildings). Furthermore, the combined objective function of smoothness and path length penalizes abrupt maneuvers, generating low-jerk trajectories, reducing gimbal vibrations and improving thermal imaging accuracy.

To summarize, the enhanced TR-SCO algorithm optimizes trajectories prioritized coverage of pedestrian-dense areas, offering a powerful technical solution for urban public health monitoring and enabling rapid identification of infected individuals to contain disease spread.
Although our proposed algorithm performed well in simulations, its application in real-world dynamic environments, such as changing weather and real-time pedestrian flow changes, needs further validation to enhance its practicality and reliability.

\section{Conclusion}
This paper introduced an enhanced Trust Region Sequential Convex Optimization algorithm specifically tailored for multi-drone thermal screening trajectory planning in complex urban environments. By addressing critical challenges such as obstacle avoidance, inter-drone coordination, trajectory smoothness, and optimal thermal imaging coverage, our proposed approach demonstrated significant improvements over conventional optimization methods. We developed a detailed simulation environment incorporating realistic urban models, drone operational constraints, and human representations to rigorously evaluate algorithm performance. Simulation experiments indicated that our enhanced TR-SCO algorithm achieved more efficient trajectory planning, reduced computational complexity, and improved thermal screening effectiveness. These advancements present valuable contributions to drone-based public health monitoring, particularly in rapid and accurate detection of elevated human body temperatures within densely populated urban areas. Future research may focus on integrating dynamic environmental conditions, real-world validation, and extending this optimization framework to accommodate larger drone fleets and more complex operational scenarios.

\bibliographystyle{elsarticle-num}
\bibliography{sample}

\end{document}